\def\year{2021}\relax
\documentclass[letterpaper]{article} 
\usepackage{amsmath}
\usepackage{amssymb}
\usepackage{comment}
\usepackage{subcaption}
\usepackage{aaai22}  
\usepackage{times}  
\usepackage{helvet}  
\usepackage{courier}  
\usepackage[hyphens]{url}  
\usepackage{graphicx} 
\usepackage{natbib}  
\usepackage{caption} 
\DeclareCaptionStyle{ruled}{labelfont=normalfont,labelsep=colon,strut=off} 
\usepackage{amsmath}

\title{Transfer Reinforcement Learning for Differing Action Spaces via Q-Network Representations}
\author{
    Nathan Beck,
    Abhiramon Rajasekharan,
    Hieu Tran
}
\affiliations{
    
    \{nathan.beck, abhiramon.rajasekharan, trunghieu.tran\}@utdallas.edu\\
    Department of Computer Science,\\
    University of Texas at Dallas,
    Richardson, TX 75080
}

\begin{document}
\maketitle

\begin{abstract}
Transfer learning approaches in reinforcement learning aim to assist agents in learning their target domains by leveraging the knowledge learned from other agents that have been trained on similar source domains. For example, recent research focus within this space has been placed on knowledge transfer between tasks that have different transition dynamics and reward functions; however, little focus has been placed on knowledge transfer between tasks that have different action spaces. In this paper, we approach the task of transfer learning between domains that differ in action spaces. We present a reward shaping method based on source embedding similarity that is applicable to domains with both discrete and continuous action spaces. The efficacy of our approach is evaluated on transfer to restricted action spaces in the Acrobot-v1 and Pendulum-v0 domains~\cite{gym}. A comparison with two baselines shows that our method does not outperform these baselines in these continuous action spaces but does show an improvement in these discrete action spaces. We conclude our analysis with future directions for this work.

\end{abstract}

\section{Introduction}

Reinforcement learning (RL) is a proven methodology for solving sequential decision-making tasks; however, as RL methods become more capable in addressing these tasks, they also tend to incur large computational costs before reaching a desired level of performance. Interestingly, many task domains bear resemblance to one another, which has prompted the study of leveraging previous RL solutions for old tasks in solving new, related tasks. This approach in RL is called \emph{transfer learning} as the knowledge learned by RL agents in older source domains is transferred to an RL agent to assist in learning a new domain. While transfer learning has been studied extensively in other areas of machine learning, it admits a promising direction for improving the performance of RL agents on increasingly difficult tasks.

A variety of transfer learning approaches exist in RL. Of the transfer learning approaches mentioned in~\cite{survey}, many choose to transfer knowledge between domains by using incentives via the reward function (known as \emph{reward shaping}) or by encoding knowledge within embedding spaces (known as \emph{representation transfer}). In particular, the advent of deep RL has encouraged heavier use of representation transfer due to the expressive power of deep neural networks~\cite{survey}. For example,~\cite{dynamicsreward} choose to use LSTM networks to learn embedding spaces that help transfer knowledge across domains differing in transition dynamics and rewards; similarly,~\cite{skillembedding} learn latent spaces that facilitate specialization in new domains that differ in transition dynamics and goal states. While such methods help facilitate knowledge transfer across domains that differ by these aspects, not as much focus has been placed on transfer across domains that differ in their action spaces, which presents a roadblock for facilitating knowledge transfer across a wide number of domains.

In this paper, we take a step towards accomplishing knowledge transfer across domains that differ only in their action spaces. Specifically, we propose a method of knowledge transfer that combines the functionality of reward shaping and representation transfer by calculating reward incentives via similarity measures of a learned embedding space. Our method can be applied to transfer settings that employ discrete action spaces or continuous action spaces. To analyze this method, we conduct experiments in the Acrobot-v1 and Pendulum-v0 domains~\cite{gym} by transferring knowledge from each to similar versions of themselves that feature restricted action spaces. While our method does not achieve effective knowledge transfer in continuous action spaces, we show that our method does achieve a measure of knowledge transfer in discrete action spaces. We conclude by discussing future directions for this work, contributing possible settings wherein our method can be expanded.

\section{Related Work}

\begin{figure*}
    \centering
    \includegraphics[width=\linewidth,height=5.5cm]{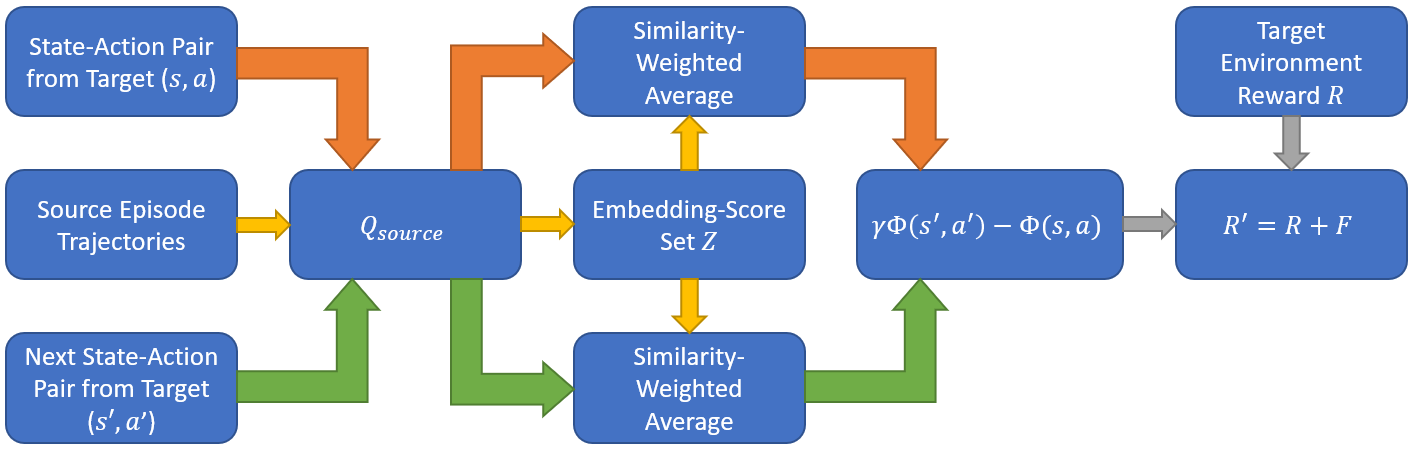}
    \caption{Summary of our reward shaping algorithm. $Q_{source}$ is used to predict the embeddings for the state-action pairs from the source trajectories and the target domain. These embeddings are then used to compute a similarity-weighted potential that is then used to provide the auxiliary reward.}
    \label{fig:alg}
\end{figure*} 

RL research in transfer learning often focuses on transfer settings where a target environment is a perturbation of the source environment(s) in some aspect.~\cite{sf_transfer} introduce an approach that learns faster in simulated navigation and robotic arm settings with modified reward functions. The approach decouples the transition dynamics and reward function from the environment so that changes in the reward function can instead be accommodated as changes in the weights associated with successor features. In a similar vein,~\cite{skillembedding} and~\cite{dynamicsreward} use decoupled dynamics and reward functions to adapt to changes in both these aspects.~\cite{skillembedding} formulate the problem as variational inference, where they learn the policy over a latent space that contains what they refer to as skill embeddings. These skill embeddings are then leveraged as opposed to using the transition dynamics directly.~\cite{dynamicsreward} use an encoder-decoder approach to learn a latent representation of the state space. A separate rewards module that uses these latent embeddings is then learned, allowing one to transfer knowledge across domains by simply interchanging the modules for shifts in the dynamics or reward function. Our research characteristically differs from these approaches as we tackle transfer between domains with different action spaces, which requires other kinds of representations than what has been previously studied. 

In related reward shaping research,~\cite{pbrs} introduce a potential-based reward shaping method for transferring domain knowledge. They show that injecting domain-based heuristics via their potential function can reduce learning time while keeping the optimal policy unchanged. Extending this idea,~\cite{pba} propose an approach where the potential function can be designed using both states and actions, which extends the state-only potential function proposed in~\cite{pbrs}. Instead of the manually designed heuristics discussed in these two works,~\cite{policy_transfer} use a policy-based reward shaping method to hasten learning. Using a learned policy in a source domain,~\cite{policy_transfer} use this approach to transfer to a target domain with different transition dynamics. They compute the auxiliary reward using the probability of a mapped state-action pair's occurrence in the source domain's learned policy. Our approach utilizes the potential-based approach presented in~\cite{pba}; however, it differs in the manner by which we compute the auxiliary reward as we tailor the computation for transfer between domains with different action spaces.

\section{Method}

\begin{table*}
\centering
\begin{tabular}
{ |l|l||l|l||l|l| }
 \hline
 \multicolumn{2}{|c|}{Source Environments} 
 & \multicolumn{2}{|c|}{Target Environments} 
 & \multicolumn{1}{|c|}{}
 & \multicolumn{1}{|c|}{}\\
 \hline
 \multicolumn{1}{|c|}{Name} & \multicolumn{1}{|c|}{Action Space} & \multicolumn{1}{|c|}{Name} & \multicolumn{1}{|c|}{Action Space} & \multicolumn{1}{|c|}{Type} & \multicolumn{1}{|c|}{Algorithms}\\
 \hline
  Pendulum~\cite{gym} & [low, high]  & New Pendulum & [0, high] & Box & TD3, DDPG \\
  Acrobot~\cite{acrobot} & Discrete(3) & New Acrobot & Discrete(2) & Discrete & DQN \\
 \hline
\end{tabular}
\caption{Characteristics of each environment and how their action spaces differ. In the case of Acrobot~\cite{acrobot}, the "no-torque" option is removed in the target space.}
\label{table:EnvironmentTabelInfo}
\end{table*}

We adopt the definition of an MDP given in~\cite{survey}. Let $\mathcal{M}_{source} = (S,A_{source},R,T,\gamma,\mu_0,S_0)$ denote the source MDP, where $S$ denotes the state space, $A_s$ denotes the action space of the source domain, $R:S \times A \times S \rightarrow \mathbb{R}$ denotes the reward function, $T:S \times A \times S \rightarrow \mathbb{R}$ denotes the transition function, $\gamma \in (0,1]$ denotes the discount factor, $\mu_0$ denotes the set of initial states, and $S_0$ denotes the set of absorbing states. Let $\mathcal{M}_{target} = (S,A_{target},R,T,\gamma,\mu_0,S_0)$ denote the target MDP, where $A_{target}$ denotes the action space of the target domain. Reinforcement learning methods can choose to learn the unknown elements of each MDP explicitly to elicit an optimal policy; otherwise, reinforcement learning methods typically derive policies based on direct estimates of a value function. Here, we are interested in instantiations of the latter case involving the use of deep neural networks for the Q-functions. Let $Q_{source}:S\rightarrow \mathbb{R}^{|A_{source}|}$ denote the Q-function estimator for a source domain with a discrete action space. In this work, we assume that the neural network architecture for $Q_{source}$ employs a fully connected layer $l$ immediately before the final output such that $Q_{source}(s) = l(Q^{last}_{source}(s))$. Let $z(s)=Q^{last}_{source}(s)$ be the features of the last linear layer for input $s$. Without loss of generality, we define $Q_{source}:S\times A_{source} \rightarrow \mathbb{R}$ as the Q-function estimator for a source domain with a continuous action space for actor-critic methods~\cite{actorcritic}.

To employ knowledge transfer from $\mathcal{M}_{source}$ to $\mathcal{M}_{target}$, we use a reward shaping approach~\cite{survey,pbrs} based on the Potential-Based State-Action Advice method given in~\cite{pba}. Namely, the target reward function $R_{target}$ is augmented with an auxiliary reward:
\begin{align*}
    R' &= R + F\\
    &= R + (\gamma \Phi(s',a') - \Phi(s,a))
\end{align*}
\noindent where $s'$ denotes the resulting state of performing action $a$ in state $s$ and $a'$ denotes the action to take by the policy when in state $s'$. By carefully designing the potential function $\Phi$, one can effectively transfer knowledge from the source domain to the target domain by incentivizing the target agent with rewards from the source agent. Hence, a target agent learns in the modified target MDP $\mathcal{M}'_{target} = (S,A_{target},R',T,\gamma,\mu_0,S_0)$. In this work, we choose to directly use the learned Q-function from $\mathcal{M}'_{target}$ when calculating the policy to use for $\mathcal{M}_{target}$ as we also explore the use of actor-critic methods~\cite{actorcritic}, which already provide the policy in the form of the actor. Note, however, that~\cite{pba} provide a method for eliciting the optimal policy for $\mathcal{M}_{target}$ given $\Phi$ and $Q^*$, the optimal Q-function for $\mathcal{M}'_{target}$, when using simpler methods.

We compute $\Phi(s,a)$ via a similarity-based average of the Q-values seen in a source episode's trajectory. Namely, we first sample $n$ episodes of a source agent and calculate the Q-network embedding $z(s)$ ($z(s,a)$ for actor-critic methods) for each state-action pair seen in these episodes. The Q-value corresponding to each state-action ($Q_{source}$) is also calculated. We store each tuple $(e_i,q_i)$ in $Z$, a set containing each embedding-value pair. Computing $\Phi(s,a)$ for a state-action pair in the target domain is then done via the following equation:
\begin{align}
    \Phi(s,a) = \frac{1}{|Z|} \sum_{i=1}^{|Z|} \frac{<z(s), e_i>}{||z(s)|| \cdot ||e_i||}q_i
\end{align}
\noindent Hence, larger potentials are given to state-action pairs in the target domain that match high Q-value state-action pairs seen in the source domain (by using cosine similarity). Note that this calculation for $\Phi(s,a)$ allows one to meaningfully calculate an auxiliary value for a state-action pair of the target domain, even if the action is not present in the source domain. For discrete action spaces, the embedding of the target state-action pair only depends on the state; for continuous action spaces, the source Q-network can still accept real-valued inputs for actions that it may not have seen in the source domain. 

Our approach is summarized in Figure~\ref{fig:alg}. As noted in~\cite{survey}, reward shaping approaches are often desirable as they are minimally intrusive to the underlying reinforcement learning algorithm. Accordingly, this allows our approach to be applied to multiple reinforcement learning algorithms with little difficulty. We utilize this benefit to adapt reinforcement learning algorithms to different kinds of action space transfer. Namely, we apply our reshaping approach to the Deep Deterministic Policy Gradient (DDPG) method~\cite{ddpg} and the Twin Delayed Deep Deterministic Policy Gradient (TD3) method~\cite{td3} to study knowledge transfer across differing continuous action spaces. Likewise, we apply our reshaping approach to a simple Deep Q-Network (DQN) method~\cite{dqn} to study knowledge transfer across differing discrete action spaces.

\section{Experiments}

\begin{figure*}
    \centering
    \includegraphics[width=0.8\linewidth]{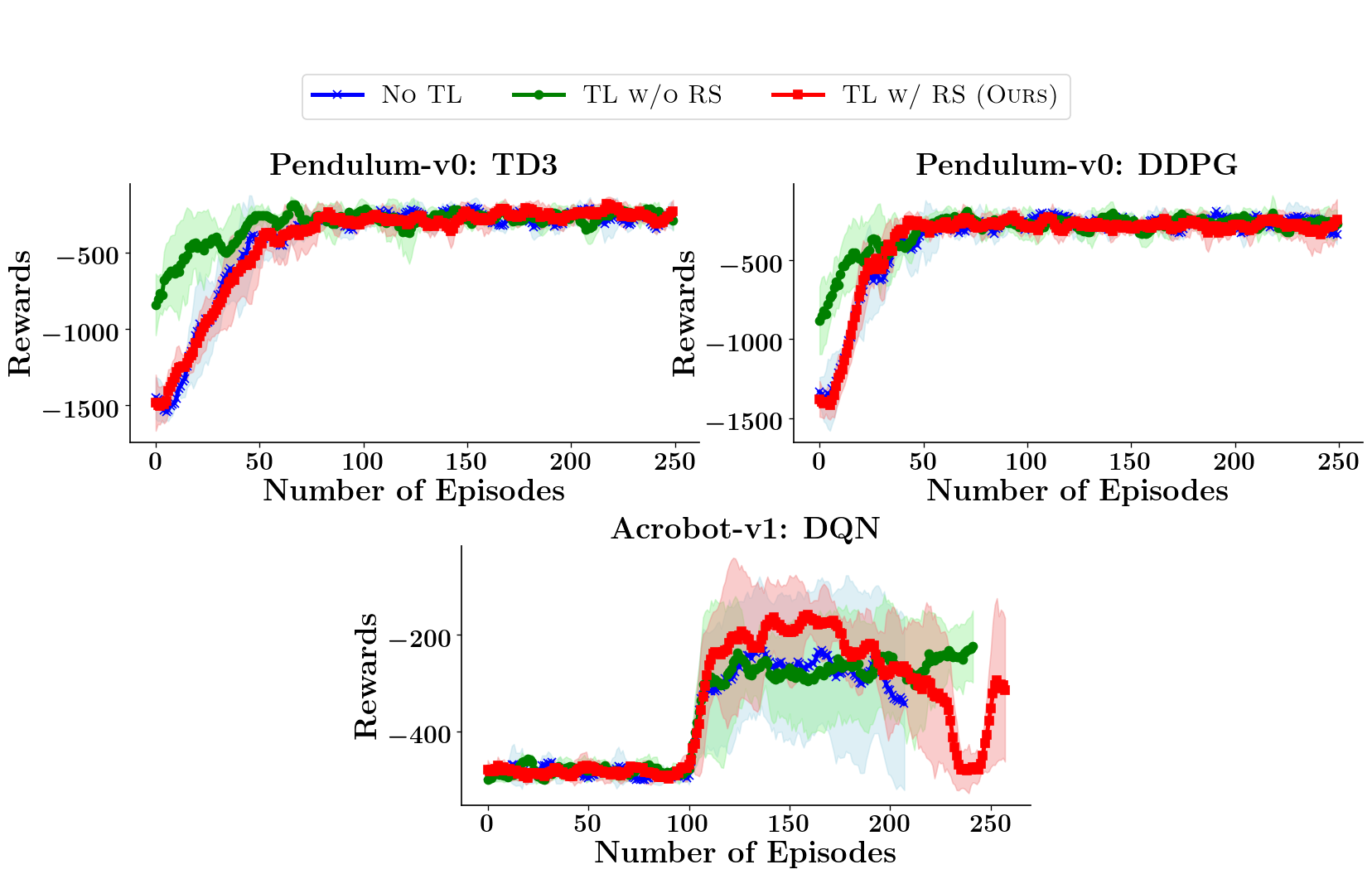}
    \caption{Experiment results conducted on Pendulum-v0 and Acrobot-v1 domains~\cite{gym}. The continuous Pendulum-v0 setting suggests that our method does not achieve significant transfer in restricted continuous action spaces. The discrete Acrobot-v1 setting suggests that our method does help in reward convergence after the initial replay buffer generation.}
    \label{fig:results}
\end{figure*}

\textbf{Implementation details}. We utilize Stable Baselines3 (SB3)~\cite{stablebaselines3}, which provides open-source implementations of reinforcement learning algorithms in PyTorch. The framework is easily modifiable by users, and it provides a clean and simple interface, giving us access to off-the-shelf RL algorithms. Accordingly, our reward shaping approach and baseline comparisons are implemented via a modified SB3\footnote{https://github.com/saodem74/Transfer-Learning-in-Reinforcement-Learning}. 

\textbf{Experimental settings}. OpenAI Gym~\cite{gym} is a popular toolkit for comparing and evaluating reinforcement learning algorithms. The Gym library provides is a collection of test environments that are reusable, modular, and modifiable via a collection of shared interfaces. As such, we use this functionality to create modifications of the Pendulum-v0 and Acrobot-v1 environments~\cite{gym,acrobot} that feature restricted action spaces. These modified environments are the target environments in our experiments; the original versions of each environment are used as source environments. Once a source model has been trained for each environment, our approach samples ten trajectories as set $Z$.

Action spaces in the Gym library can be classified into four groups: a) \textsc{Box}, a $N$-dimensional box in $\mathbb{R}^N$ whose tuples represent a possible action; b) \textsc{Discrete}, a list of possible actions where only one of the actions can be used per timestep; c) \textsc{MultiDiscrete}, a list of action sets where only one action of each discrete set can be used per timestep; d) \textsc{MultiBinary}, a list of possible actions where any of the actions can be used in any combination per timestep. In our scope, we consider only the \textsc{Discrete} and \textsc{Box} action spaces to handle both discrete and continuous action spaces, respectively. In the \textsc{Discrete} space, new environments are created by taking a subset of the original environment's action space. In the \textsc{Box} space, we clip the action values so that the new action lies within a subset of the original box. The details of each environment are demonstrated in Table~\ref{table:EnvironmentTabelInfo}.

\textbf{Model comparison}.
We evaluate our approach's episodic reward performance in comparison to two other baselines. The first baseline learns the target domain from scratch without any transferred knowledge from source domain. The second baseline directly applies the pre-trained source model in the target domain, necessitating the adaptation of the source model to the target domain. Each method that uses knowledge transfer receives the same source model.

\textbf{Experimental results}.
All experiments were run on Google Colab, which offers runtimes with V100 and P100 GPUs. We run experiments across both environments using the previously mentioned RL methods where they are applicable: DDPG~\cite{ddpg}, TD3~\cite{td3}, and DQN~\cite{dqn}. For each method-environment combination, the three approaches are conducted, and their results are presented in Figure~\ref{fig:results}. Each plot presents the learning curve of the no-transfer baseline (blue), direct-transfer baseline (green), and our reward shaping approach (red). The models in each approach are trained up to 50k timesteps for the Pendulum-v0 environment and 100k timesteps for the Acrobot-v1 environment. After obtaining the episode rewards, we smooth the plots by applying a moving average with window size $W=7$. We repeat this procedure for a total of $n=5$ times, producing the error bars given in the plot.

As can be seen in Figure~\ref{fig:results}, our approach performs only as well as the no-transfer baseline in the continuous setting, being only marginally better for a few episodes in DDPG. For restricted action spaces in the continuous setting, the results indicate that clipping the action output of the source model is sufficient. In the discrete setting, Figure~\ref{fig:results} also demonstrates the results in the New Acrobot environment with the underlying DQN method. While the error bars for these experiments are large, our reward shaping approach appears to converge faster than these two baselines on average as it converges to a higher reward after the 100 initial episodes of populating the replay buffer with trajectories. Surprisingly, the learning curve for our reward shaping method fluctuates drastically after 200 episodes. We suspect that this behavior is characteristic of the environment; however, additional experimentation would be needed to confirm this observation.

\section{Discussion}

As discussed before, our method achieves some measure of knowledge transfer to restricted discrete action spaces. However, what remains unexplored in this thread is the application of this method for transferring knowledge to \emph{expanded} action spaces and action spaces that have disjoint components. For target domains that feature the discrete portion of these action spaces, our method can be still be applied since the input for the Q-networks used in these discrete spaces only depend on the state. The same observation applies to the continuous portion of these target domains since the corresponding critic networks already accept inputs from $\mathbb{R}^n$. Another unexplored direction is the application of our method to other underlying RL algorithms. While we study the use of DQNs~\cite{dqn}, DDPG~\cite{ddpg}, and TD3~\cite{td3}, the analysis of our method applied to other recent RL frameworks that employ Q-networks would provide further knowledge as to the efficacy of our method. In summary, our method presents a plausible direction for performing knowledge transfer across differing action spaces.

\bibliography{main}

\end{document}